\documentclass{article}
\usepackage[T1]{fontenc}
\usepackage[authoryear,round]{natbib}
\usepackage[preprint]{neurips_2025_custom}

\usepackage{amsmath,amsfonts,bm}

\def\eqref#1{equation~\ref{#1}}

\def\1{\bm{1}}

\DeclareMathAlphabet{\mathsfit}{\encodingdefault}{\sfdefault}{m}{sl}
\SetMathAlphabet{\mathsfit}{bold}{\encodingdefault}{\sfdefault}{bx}{n}

\usepackage{array,booktabs,longtable,multirow,graphicx,xspace,xcolor}
\usepackage{hyperref,url}
\hypersetup{hidelinks}
\definecolor{linkColor}{rgb}{0.18,0.39,0.62}
\usepackage{colortbl}
\usepackage{pifont,makecell}
\usepackage{enumitem}
\usepackage{listings}
\usepackage{needspace}
\usepackage{float,flafter}
\usepackage{caption,adjustbox,placeins}

\newcommand{\cmark}{\textcolor{green!60!black}{\ding{51}}}
\newcommand{\xmark}{\textcolor{red!80!black}{\ding{55}}}

\definecolor{baselinegray}{RGB}{240,240,240}
\definecolor{oursgreen}{RGB}{220,242,230}
\definecolor{ourspink}{RGB}{252,232,238}

\usepackage{xcolor}

\definecolor{PaperRed}{RGB}{255,136,136}      %
\definecolor{PaperOrange}{RGB}{255,170,68}    %
\definecolor{PaperYellow}{RGB}{255,204,68}    %
\definecolor{PaperGreen}{RGB}{102,238,102}    %
\definecolor{PaperBlue}{RGB}{136,204,255}     %
\definecolor{PaperPurple}{RGB}{204,136,255}   %

\colorlet{PaperRedLight}{PaperRed!50!white}
\colorlet{PaperOrangeLight}{PaperOrange!50!white}
\colorlet{PaperYellowLight}{PaperYellow!50!white}
\colorlet{PaperGreenLight}{PaperGreen!50!white}
\colorlet{PaperBlueLight}{PaperBlue!50!white}
\colorlet{PaperPurpleLight}{PaperPurple!50!white}

\definecolor{medicalgray}{RGB}{252,246,224}

\definecolor{oursorange}{RGB}{253,240,228}
\colorlet{oursyellow}{PaperYellowLight}

\definecolor{MedalGold}{HTML}{B99A45}
\definecolor{MedalSilver}{HTML}{B0B0B0}
\definecolor{MedalBronze}{HTML}{8C593B}
\DeclareRobustCommand{\rankmedal}[1]{\textcolor{#1}{\ensuremath{\bullet}}}
\newcommand{\rankedscore}[2]{#2\rlap{\hspace{0.15em}\rankmedal{#1}}}
\newcommand{\best}[1]{\rankedscore{MedalGold}{#1}}
\newcommand{\second}[1]{\rankedscore{MedalSilver}{#1}}
\newcommand{\third}[1]{\rankedscore{MedalBronze}{#1}}

\newcommand{\compacttable}{%
  \footnotesize
  \setlength{\tabcolsep}{4pt}%
  \renewcommand{\arraystretch}{1.0}%
}

\newcolumntype{P}[1]{>{\raggedright\arraybackslash}p{#1}}
\newcommand{\method}{TUNE\xspace}

\newcommand{\papertitle}{Harnessing Medical Visual-Language Models with Agentic Evidence Management}
\title{\fontsize{15.5pt}{18pt}\selectfont\papertitle}
\author{%
\begin{tabular}{c}
\\[2pt]
\textbf{Huang Qiming}\textsuperscript{1}\thanks{Equal contribution.\quad\textsuperscript{\textdagger}\,Corresponding author.}
\quad
\textbf{Hanze Dong}\textsuperscript{2}\footnotemark[1]
\quad
\textbf{Jipeng Zhang}\textsuperscript{2}\footnotemark[1]\\[2pt]
\textbf{Mike Zheng Shou}\textsuperscript{1}
\quad
\textbf{Xinxing Xu}\textsuperscript{2}
\quad
\textbf{Furu Wei}\textsuperscript{2\textdagger}\\[2pt]
\textnormal{\textsuperscript{1}National University of Singapore}\\
\textnormal{\textsuperscript{2}Microsoft}\\
\textcolor{linkColor}{\url{https://aka.ms/GeneralAI}}
\end{tabular}
}
\AtBeginDocument{\setlength{\headheight}{14.4pt}}
\makeatletter
\renewcommand{\@noticestring}{Technical report.}
\makeatother
\begin{document}
\maketitle
\setcounter{footnote}{0}
\vspace{-30pt}
\begin{abstract}
Medical vision--language agents increasingly rely on segmentation tools, and recent work has focused on acquiring the right mask, often treating a good mask as the end point of tool use.
However, do these tool-augmented agents actually make effective use of the evidence once it is obtained?
To answer this question, we need a benchmark that separates evidence use from segmentation quality, spans multiple medical image types, and allows the same mask to be presented in different ways.
Existing medical VQA datasets do not jointly provide these properties.
We therefore construct \textbf{Mask-Evidence MedVQA}, a 2.5K-example benchmark spanning four image types, where each question is paired with an answer-relevant ground-truth mask and multiple possible renderings.
Our study reveals two limitations of direct tool use.
First, \textbf{evidence presentation} matters. Different renderings of the same fixed mask can substantially change model behavior.
Second, masks introduce an \textbf{evidence trade-off}. Highlighting a relevant region can expose diagnostic evidence the model previously missed, but the added visual cue can also induce incorrect associations. As a result, even a ground-truth mask can correct some answers while turning initially correct ones into incorrect ones.
Together, these findings show that acquiring the right mask is not enough; an agent must also decide how to present the evidence and whether it should change the answer.
We therefore introduce \mbox{\textbf{TUNE} (\textbf{T}ool \textbf{U}se is \textbf{N}ot \textbf{E}nough)}, an evidence-management harness that explicitly manages what evidence to acquire, how to present it, and whether to accept the resulting answer update.
Across three public medical VQA benchmarks, evidence management improves the fine-tuned Qwen3.5-9B backbone's mean accuracy by 4.0 percentage points. Full TUNE improves over direct tool use by 9.1 points on average across seven benchmarks and achieves competitive results against existing medical VLMs and agents.

\end{abstract}
\section{Introduction}
\label{sec:intro}

Medical vision--language model (VLM) agents increasingly consult specialist tools before answering questions about medical images~\citep{mmedagent,medrax,care,ophiuchus,mira}.
Segmentation models such as MedSAM and BiomedParse are particularly useful because they return an explicit mask of the region relevant to the question~\citep{medsam,biomedparse}.
Accordingly, recent work has focused on improving evidence acquisition: deciding when to invoke a tool, what finding or structure to segment, and how to obtain the relevant region reliably~\citep{ophiuchus,care,mira}.
The field has thus improved mask acquisition while largely treating the returned mask as the end point of tool use. But do these tool-augmented agents actually make effective use of the evidence once it is obtained?

Answering this question requires \textbf{multiple medical image types} to test generality across domains.
It also requires \textbf{masks as additional evidence}, so no-tool answers exist and mask-induced changes can be measured.
Finally, \textbf{multiple renderings per mask} are needed to vary presentation while holding localization fixed.
Existing medical VQA benchmarks provide at most two of the three, as shown in Table~\ref{tab:dataset_comparison}.
We therefore construct Mask-Evidence MedVQA, with 2,494 questions across ultrasound, chest X-ray, endoscopy, and CT, each paired with a relevant ground-truth mask.
Each mask can be rendered as an outline, calipers, a crop, an inset, or another style used in medical figures.
Using ground-truth masks factors mask quality out of tool-use analysis, allowing us to measure how evidence use varies across image types and renderings.

\begin{figure}[t]
\centering
\includegraphics[width=1.0\linewidth]{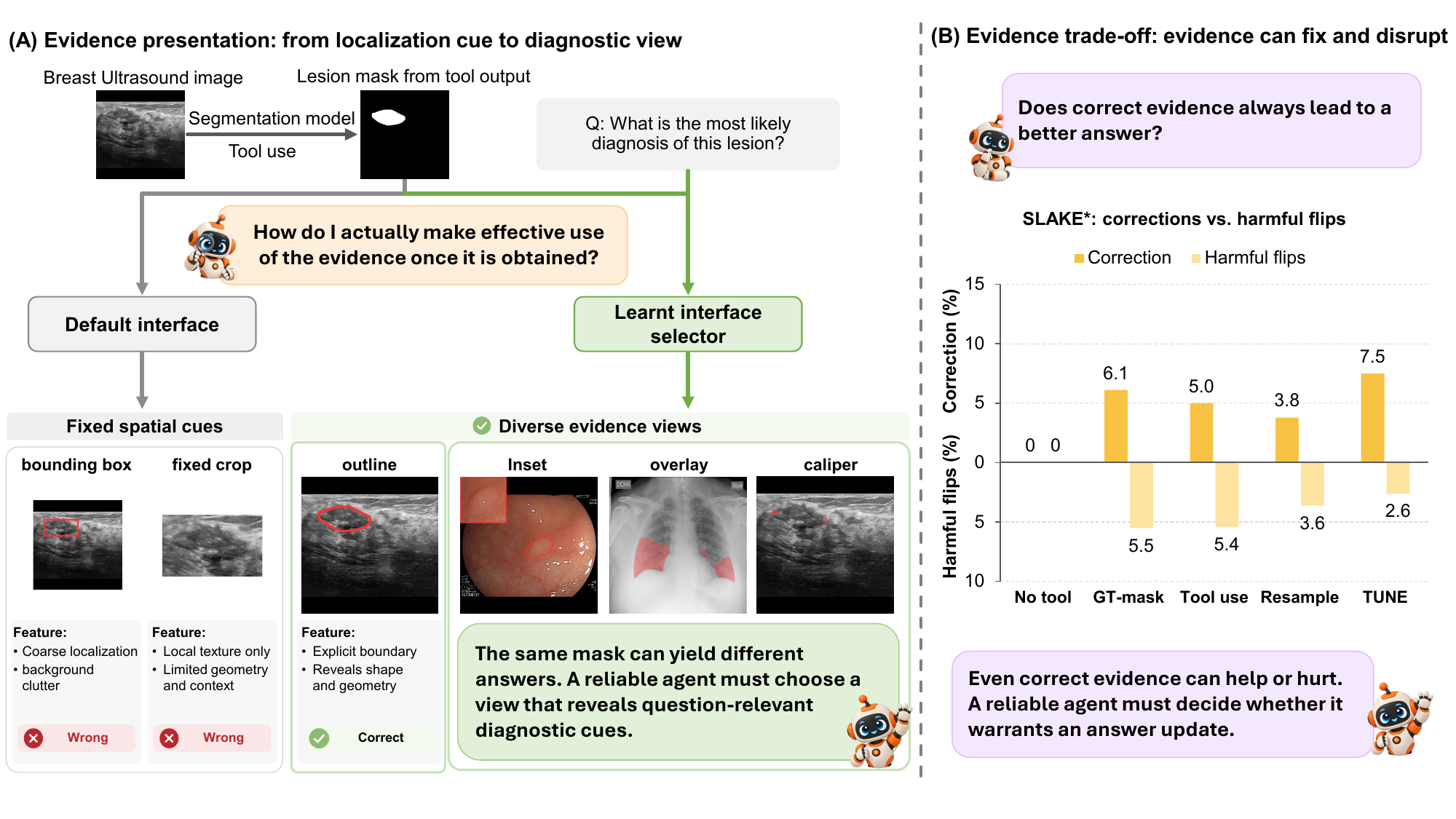}
\caption{
\textbf{Tool use is not enough: reliable evidence use requires choosing how to present a mask and whether to update the answer.}
\textbf{(A) Evidence presentation.}
The same mask can lead to different predictions depending on its rendering.
In the ultrasound example, bounding-box and fixed-crop views yield incorrect answers, whereas an outline exposes lesion shape and yields a correct answer.
Adaptive presentation selects among complementary views to reveal question-relevant diagnostic cues.
\textbf{(B) Evidence trade-off.}
On SLAKE$^{*}$, upward bars show corrections (incorrect to correct) and
downward bars show harmful flips (correct to incorrect), both relative to
the SFT no-tool answers. Among the settings that use masks, TUNE achieves
the highest correction rate (7.5\%) and the lowest harmful-flip rate (2.6\%).
}
\label{fig:teaser}
\end{figure}

First, the same localized evidence can lead to different answers depending on how it is presented.
In Fig.~\ref{fig:teaser}(A), the underlying ground-truth mask is identical, but different renderings, such as a crop or an outline, yield different predictions.
This suggests that the model does not use a mask merely as a localization cue.
An outline emphasizes boundary shape, calipers emphasize size, and a crop or inset changes the balance between local detail and surrounding context.
Evidence presentation is thus part of the reasoning process: holding localization fixed does not hold the effective evidence seen by the model fixed.
This interpretation is consistent with prior findings on visual prompting and the diagnostic conventions used in medical figures~\citep{redcircle,bioimage_figures}.

Second, correct evidence is not uniformly beneficial.
Figure~\ref{fig:teaser}(B) shows that even ground-truth masks can correct
initially wrong answers and overturn initially correct ones on SLAKE$^{*}$.
We observe the same trade-off across Mask-Evidence MedVQA.
Highlighting can make question-relevant evidence easier to access, but it also changes visual salience and context and may introduce presentation-specific shortcuts.
This exposes an evidence trade-off between informativeness and presentation-induced interference.
Making relevant evidence more explicit can help the model discover missed information, but can also disrupt an otherwise correct prediction.
Such failures are consistent with prior observations on visual cue-following and spurious correlations in medical imaging~\citep{redcircle,jabbour2020shortcuts}.
Therefore, a reliable agent should compare the evidence-conditioned answer with its original answer and decide whether the new evidence warrants an update.

These findings suggest that reliable tool use requires managing what evidence is acquired, how it is presented, and whether it should change the answer. We therefore introduce \textbf{TUNE} (\textbf{T}ool \textbf{U}se is \textbf{N}ot \textbf{E}nough), a harness that makes these decisions explicit from question to final prediction.
TUNE consists of three stages. Building on prior work in evidence acquisition, the \textbf{Target Planner} converts the question into an executable request for answer-relevant visual evidence. To address the effects of presentation, the \textbf{Evidence Presenter} selects how the returned mask should be rendered. Because even correct evidence can hurt, the \textbf{Answer Critic} compares the original and evidence-conditioned predictions and decides whether the new evidence warrants an update.
Together, these stages turn one-shot ``retrieve-and-answer'' tool use into explicit evidence management, deciding what evidence to request, how to communicate it, and whether to act on it.
We further co-evolve TUNE and the agent through two stages of harness refinement and model adaptation. The first refined harness generates evidence-management trajectories to update the model. The updated model then informs a second harness refinement, whose trajectories support final model training.

We evaluate TUNE on three public medical VQA benchmarks, SLAKE, VQA-RAD, and OMVQA-3k~\citep{slake,vqarad,omnimedvqa}, and the four datasets of Mask-Evidence MedVQA.
Together, these datasets span nine image modalities and multiple organs.
Evidence management improves the fine-tuned Qwen3.5-9B backbone by 4.0 percentage points on average across the three public benchmarks. Full TUNE improves over direct tool use by 9.1 points across all seven datasets and achieves competitive results against existing medical VLMs and agents.
In summary, our main contributions are:
\begin{itemize}[leftmargin=*,itemsep=0pt,topsep=1pt]
\item \textbf{A new benchmark.} Mask-Evidence MedVQA combines multiple medical image types, ground-truth masks as evidence, and multiple renderings per mask to measure evidence use while factoring out mask quality.

\item \textbf{An evidence-management harness.} We introduce TUNE to manage what evidence to acquire, how to present it, and whether that evidence warrants an answer update.

\item \textbf{Co-evolutionary training.} We co-evolve TUNE and the agent: trajectories from a refined harness support model adaptation, while the updated model informs further harness refinement before final training.

\item \textbf{Competitive results.} TUNE improves over direct tool use on all seven datasets and achieves competitive performance on public medical VQA benchmarks.
\end{itemize}

\section{Related Work}
\label{sec:related_work}

\paragraph{Medical visual-tool agents and harnesses.}
Medical VLM agents increasingly call specialist tools before answering~\citep{mmedagent,medrax,care,ophiuchus}, while foundation segmenters such as MedSAM and BiomedParse identify explicit image regions across modalities~\citep{medsam,biomedparse}.
Recent systems improve this process in different ways.
CARE proposes entities to segment and reviews evidence-grounded answers; Ophiuchus learns when and where to inspect~\citep{care,ophiuchus}.
MIRA searches over image operations, while Med-VRAgent coordinates region-level reasoning~\citep{mira,medvragent}.
These approaches largely focus on obtaining relevant evidence.
We instead treat acquisition as the first stage of \emph{evidence management}.
Subsequent stages decide how to present the returned evidence and whether it should change the final answer.

\begin{table}[!t]
\centering
\hypersetup{hidelinks}
\caption{\textbf{Medical VQA benchmarks and support for evidence evaluation.}
\# QA denotes the number of question--answer pairs; image types denote imaging modalities.
Mask as evidence means the original image alone supports an answer, with the mask
providing additional evidence. Multiple renderings are alternative views of the same mask.}
\label{tab:dataset_comparison}

\small
\setlength{\tabcolsep}{5.5pt}
\renewcommand{\arraystretch}{0.92}

\begin{adjustbox}{max width=0.95\linewidth}
\begin{tabular}{lcccc}
\toprule
\multicolumn{1}{c}{\makecell[c]{\textbf{Dataset}}}
& \makecell[c]{\textbf{\# QA}}
& \makecell[c]{\textbf{Multiple}\\\textbf{image types}}
& \makecell[c]{\textbf{Mask as}\\\textbf{evidence}}
& \makecell[c]{\textbf{Multiple}\\\textbf{renderings}} \\
\midrule
VQA-RAD~\citep{vqarad}
& 0.45K & \cmark & \xmark & \xmark \\
SLAKE~\citep{slake}
& 1.06K & \cmark & \xmark & \xmark \\
PathVQA~\citep{pathvqa}
& 6.76K & \xmark & \xmark & \xmark \\
PMC-VQA~\citep{pmcvqa}
& 2.00K & \cmark & \xmark & \xmark \\
Halt-MedVQA~\citep{haltmedvqa}
& 2.36K & \cmark & \xmark & \xmark \\
HEAL-MedVQA~\citep{healmedvqa}
& 2.85K & \xmark & \cmark & \xmark \\
\midrule
\rowcolor{oursyellow}
\textbf{Mask-Evidence MedVQA-Eval (Ours)}
& \textbf{2.49K} & \cmark & \cmark & \cmark \\
\bottomrule
\end{tabular}
\end{adjustbox}
\end{table}

\paragraph{Visual prompting and evidence presentation.}
Prior visual-prompting work mainly studies how to direct attention to a region.
MedVP compares scribbles, rectangles, and ellipses~\citep{medvp}, while radiology studies use arrows, boxes, and circles~\citep{denner2026radiology}.
Targeted Visual Prompting pairs a local crop with the full image~\citep{targetedvp}, and CARE selects among crops, masks, and global indicators~\citep{care}.
FAVP further considers masks and contours and finds that preferred styles vary across datasets~\citep{favp}.
We take a different view: presentation should determine not only \emph{where} the model looks, but also \emph{what property} of a fixed region is made salient.
Our interface library therefore spans localization, boundary and geometry, measurement, local appearance, and surrounding context.
The Evidence Presenter learns from answer outcomes which view is useful for each question.

\paragraph{Evidence bias and answer review.}
Additional evidence can improve an answer while also introducing new biases.
Language models may abandon correct predictions after receiving external context or tool outputs~\citep{situated_faithfulness,when_tools_hurt,no_worse_context}.
Multimodal models can also respond differently to equivalent content across modalities or visual markings~\citep{modality_conflict,redcircle,winkler2019markings}.
Existing medical agents typically review the evidence-conditioned answer itself~\citep{care,mira}, which does not directly reveal whether the added \mbox{evidence} improved or merely swayed the model.
TUNE instead preserves an independent no-tool answer and reviews the update by comparing it against the evidence-conditioned answer.
Unlike self-correction methods that generate a revised answer~\citep{pag_verifier}, our Critic selects only between these two completed candidates.

\section{Is Tool Use Alone Enough?}
\label{sec:evidence_gap}

Prior work has made substantial progress in acquiring relevant visual evidence, but much less is known about how a VLM uses that evidence once it is returned.
Studying this step requires multiple medical image types, masks introduced as additional evidence, and multiple renderings of each mask.
Existing medical VQA benchmarks do not jointly provide these three properties (Table~\ref{tab:dataset_comparison}).

\textbf{Benchmark setup.}
Mask-Evidence MedVQA is constructed from four medical segmentation datasets spanning breast ultrasound, chest X-ray, endoscopy, and CT~\citep{busi,covidqu,neopolyp,kits23_dataset}.
GPT-5.5 generates multiple-choice questions with shuffled options from the ground-truth mask and source label. We retain questions that are answerable from the image alone, match the label, and concern the masked region without revealing the answer or mentioning the mask. Masks are binarized and rendered without class-coded colors; COVID-QU-Ex uses lung-field masks across all three classes. Curation rules, splits, and class statistics are provided in Appendix~\ref{app:construction}.
The resulting benchmark contains 2,494 examples with ground-truth masks that support multiple renderings.
These renderings let us hold localization fixed and isolate how the VLM uses the acquired evidence.

\textbf{Human verification.}
We draw a stratified sample of 200 benchmark examples across the four image types. Three independent reviewers assess answerability from the original image, consistency with the source label, and relevance of the ground-truth mask to the question. Of the reviewed examples, 96\% satisfy all three criteria according to at least two reviewers. Image-only answerability is 97.5\% by majority judgment; inter-reviewer agreement is Fleiss\textquotesingle{} $\kappa=0.71$ (Appendix~\ref{app:construction}).

\textbf{Findings and design implications.}
Two effects emerge from this controlled setting.
First, changing only how a fixed mask is rendered can substantially change the answer, even though the underlying localized region remains identical.
This shows that a mask is not merely a localization signal: how it is presented determines which properties of the region become salient to the VLM.
Second, even a ground-truth mask can both correct an initially wrong answer and turn an initially correct one into a wrong answer.
Thus, additional evidence carries an information--bias trade-off: it can reveal useful diagnostic information while also introducing visual cues that mislead the model.

\section{\method: A Harness for Agentic Evidence Management}
\label{sec_method}

\begin{figure}[t]
    \centering
    \includegraphics[width=0.95\linewidth]{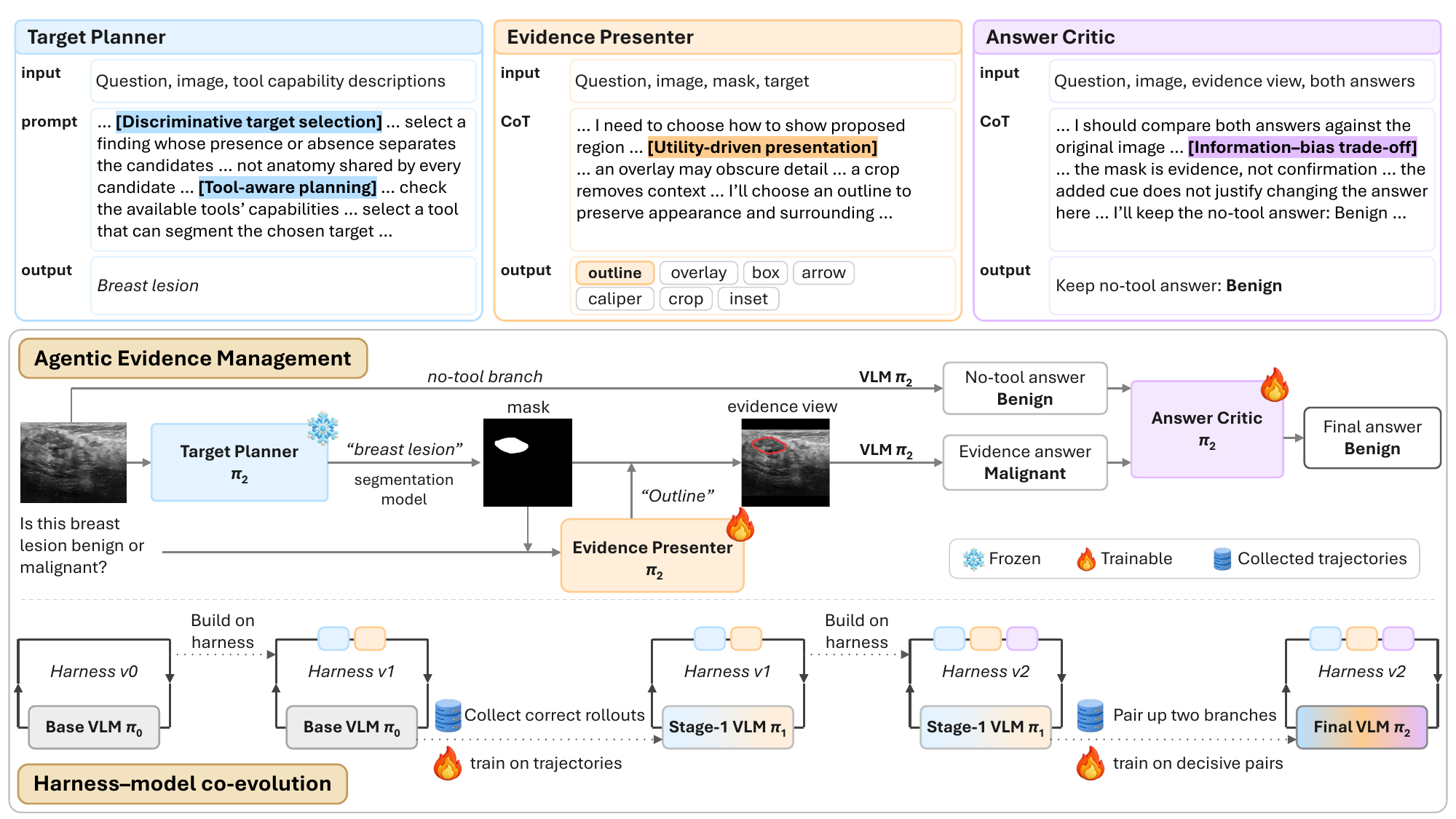}
    \caption{\textbf{TUNE manages the path from a question to a warranted answer update.}
    The top panels show the inputs and outputs of the Target Planner, Evidence Presenter, and Answer Critic.
    The middle panel traces a breast ultrasound example in which the Presenter selects an outline
    and the Critic retains the original no-tool answer after comparing both answers.
    The bottom panel shows how the harness and VLM co-evolve as harness-generated trajectories improve the VLM,
    which then uses the expanded harness to collect decisive answer pairs for the next stage of training.
    At inference time, the Target Planner, Evidence Presenter, and Answer Critic all use the same final VLM $\pi_2$ with role-specific prompts.}
    \label{fig:method}
\end{figure}

\textbf{Overview.}
Given a medical image $x$ and question $q$, \method manages segmentation evidence through three stages (Fig.~\ref{fig:method}).
The \textbf{Target Planner} decides what visual evidence to segment, and the \textbf{Evidence Presenter} selects how to present the resulting mask to the VLM.
The \textbf{Answer Critic} decides whether the evidence-conditioned answer should replace the model's original prediction.
Together, these stages manage evidence acquisition, presentation, and final answer selection.
All three roles are implemented by a shared VLM $\pi_\theta$, while the segmentation tool and image renderers remain fixed external components.

\textbf{Pipeline.}
The Planner first proposes a semantic target $t$, which is passed to a segmentation tool $S$ to obtain a mask $m$.
The Evidence Presenter chooses an interface $k$ from a fixed library $\mathcal{K}$, and a deterministic renderer produces the evidence view $v=R_k(x,m)$.
The model independently generates a no-tool answer $a_0$ from $(q,x)$ and an evidence-conditioned answer $a_1$ from $(q,x,v)$.
The Critic then compares the two candidates $(a_0, a_1)$ and selects the final answer. It is invoked only when the candidates differ; identical answers are returned directly.

\subsection{Target Planner: What Evidence to Acquire}
\label{sec_method_planner}

The Target Planner determines what evidence to acquire with the segmentation tool configured for the run.
Given the question $q$, image $x$, and the configured tool\textquotesingle{}s description, the Planner outputs a semantic target $t$ that is compatible with that tool.
The selected tool then produces a segmentation mask $m$ corresponding to the planned target.

\textbf{Discriminative target selection.}
A target that is merely relevant to the question is often too broad to provide useful evidence for reasoning.
For example, segmenting the \emph{lung} identifies a relevant anatomical region but provides little information for distinguishing among candidate pulmonary diagnoses.
Segmenting a \emph{ground-glass opacity}, by contrast, exposes evidence that may directly support or rule out candidate answers.
We therefore prompt the Planner to propose targets whose segmentation can provide \emph{discriminative evidence}, rather than simply extracting entities that are related to the question.

\textbf{Tool-aware planning.}
Segmentation models differ both in what they can segment and in the inputs they require.
BiomedParse~\citep{biomedparse} accepts text prompts for a predefined vocabulary of anatomical structures and lesions, but cannot segment targets outside that vocabulary.
MedSAM~\citep{medsam}, in contrast, takes a spatial prompt and can therefore handle targets such as \emph{teeth} once they are localized.
We describe these capabilities to the Planner and define a tool-specific workflow for each model.
BiomedParse receives the planned semantic target directly; for MedSAM, a VLM first predicts a bounding box, which MedSAM uses to segment the target.

\subsection{Evidence Presenter: How to Present the Evidence}
\label{sec_method_presenter}

The Evidence Presenter determines how the predicted segmentation mask should be presented to the VLM.
Given the question $q$, image $x$, and mask geometry, it selects an interface $k$ from the 12-element library $\mathcal{K}$ before viewing any rendering (Appendix~\ref{app:prompts}).
A deterministic renderer then produces the evidence view $v=R_k(x,m)$.
The library includes overlays, outlines, contour fills, and mask-only views for region and boundary cues, and bounding boxes, circles, crosshairs, and arrows for spatial cues.
It also includes calipers for measurement, and crop and inset views for local appearance and surrounding context.

\textbf{Utility-driven presentation.}
A mask specifies where the evidence lies, but how it is drawn shapes what the model can read from it.
An outline can expose boundary shape, and a caliper can emphasize lesion extent; a crop or inset can reveal local appearance or preserve anatomical context.
We therefore treat evidence presentation as an explicit reasoning decision rather than a fixed visualization choice.
Because medical VQA datasets do not annotate useful presentations, we learn this choice from downstream answer utility.
We hold the target and mask fixed, render every interface in $\mathcal{K}$, and supervise the Presenter using those that yield a correct evidence-conditioned answer.

\subsection{Answer Critic: Whether to Trust the Evidence}
\label{sec_method_critic}

The Answer Critic determines whether the evidence-conditioned answer should replace the model's original answer.
Before observing the rendered evidence, the VLM produces a no-tool answer $a_0$; after observing $v$, it independently produces an evidence-conditioned answer $a_1$.
The Critic receives $q$, $x$, $t$, $v$, both candidate answers, simple mask diagnostics, and relevant tool limitations, then selects the final answer.

\textbf{Information--bias trade-off.}
Targeted evidence can expose useful diagnostic information, but the added visual cue can also induce misleading associations and turn a correct prediction into an incorrect one.
We therefore preserve both the original answer $a_0$ and the evidence-conditioned answer $a_1$, and ask the Critic whether the new evidence justifies updating the model's initial judgment.
Rather than generating another answer, the Critic selects only between these two candidates.
We train it on \emph{decisive pairs}, where $a_0$ and $a_1$ disagree and exactly one is correct.
These pairs capture both cases where tool use helps and cases where it hurts.
To prevent a positional shortcut, we randomly swap the order in which the two candidates appear in the Critic prompt during both training and inference.

\subsection{Harness--Model Co-evolution}
\label{sec_method_training}

The Evidence Presenter and Answer Critic require supervision that standard medical VQA datasets do not provide.
We derive this supervision from outcome-labeled trajectories as we develop the harness and model together.

\textbf{Co-evolution recipe.}
Starting from a basic tool-use loop with the base VLM $\pi_0$, we first add the Target Planner and Evidence Presenter to improve the harness.
Using a subset of $\mathcal{D}_{\mathrm{train}}$, we enumerate the interfaces in $\mathcal{K}$ and retain rollouts with correct evidence-conditioned answers as $\mathcal{T}_0$.
Fine-tuning $\pi_0$ on $\mathcal{T}_0$ yields an intermediate model $\pi_1$ that learns to select useful evidence presentations.
We then add the Answer Critic to the harness and run $\pi_1$ over the full training set through both the no-tool and evidence-conditioned branches.
Decisive answer pairs from these runs form answer-selection trajectories $\mathcal{T}_1$, which supervise when the evidence-conditioned answer should replace the original answer.
Finally, we fine-tune the original base model $\pi_0$ on $\mathcal{T}_0 \cup \mathcal{T}_1$ to obtain $\pi_2$.
With different role prompts, this model serves as the Target Planner, Evidence Presenter, answer generator, and Answer Critic. Training details and role prompts are provided in Appendices~\ref{app:training} and~\ref{app:prompts}.

Learning when to trust evidence makes the Critic useful, so \textbf{a better harness needs a better model}. In turn, collecting answer pairs through the learned Presenter provides more useful supervision for the final model, so \textbf{a better model also needs a better harness} (Appendix~\ref{app:controls}).

\section{Experiments}
\label{sec:experiments}

We evaluate TUNE from two perspectives.
We first ask whether evidence management improves over direct tool use.
Controlled analyses and ablations then examine how evidence presentation and answer review contribute to these gains.

\subsection{Experimental Setup}
\label{sec:exp_setup}

\paragraph{Datasets.}
We evaluate on four Mask-Evidence MedVQA datasets (BreastUS, COVID-QU-Ex, NeoPolyp, and KiTS23) and three public medical VQA benchmarks (SLAKE, VQA-RAD, and OMVQA-3k)~\citep{slake,vqarad,omnimedvqa}.
Ground-truth masks are used only in controlled analyses of evidence presentation and answer changes; TUNE itself always operates on predicted masks. SLAKE* denotes test questions with an annotated mask for the queried structure. CARE-aligned comparisons use the released splits; other experiments retain our original splits.

\begin{table}[!tb]
\centering
\hypersetup{hidelinks}

\caption{
\textbf{Accuracy (\%) on public medical VQA benchmarks.}
Overall is the unweighted mean accuracy across the three datasets.
Full TUNE uses CARE's released splits and GPT-4o scoring. Prior agent results follow their published protocols.
Gold~\rankmedal{MedalGold}, silver~\rankmedal{MedalSilver}, and
bronze~\rankmedal{MedalBronze} mark the top three distinct scores per column,
with ties sharing a medal. Yellow rows denote our methods.
\method$^{\mathrm{C}}$ uses a fixed overlay; full \method includes model fine-tuning and learned evidence presentation.
}
\label{tab_benchmark_context_vqa}

\compacttable
\setlength{\tabcolsep}{3pt}

\begin{adjustbox}{max width=\linewidth}
\begin{tabular}{lc*{4}{c}}

\toprule
\multicolumn{1}{c}{\textbf{System}}
& \textbf{Access}
& \multicolumn{1}{c}{\textbf{SLAKE}}
& \multicolumn{1}{c}{\textbf{VQA-RAD}}
& \multicolumn{1}{c}{\textbf{OMVQA-3k}}
& \multicolumn{1}{c}{\textbf{Overall}} \\
\midrule

Llama-3.2-11B-Vision~\citep{llama32}
& Open
& 63.2 & 53.2 & 43.1 & 53.2 \\

Qwen3.5-9B~\citep{qwen35_9b}
& Open
& 75.1 & 69.0 & 81.4 & 75.2 \\

InternVL3-38B~\citep{internvl3}
& Open
& 68.7 & 63.0 & 78.6 & 70.1 \\

DeepEyes-7B~\citep{deepeyes}
& Open
& 61.2 & 56.1 & 57.4 & 58.2 \\

Gemma-4-E4B~\citep{gemma4}
& Open
& 69.3 & 55.4 & 70.6 & 65.1 \\

medgemma-27b~\citep{medgemma}
& Open
& 70.5 & 62.8 & 64.2 & 65.8 \\

llava-med-v1.5-mistral-7b~\citep{llavamed}
& Open
& 50.9 & 41.9 & 45.3 & 46.0 \\

MedVLM-R1-2B~\citep{medvlmr1}
& Open
& 46.5 & 41.5 & 72.1 & 53.3 \\

HuatuoGPT-Vision-34B~\citep{huatuovision}
& Open
& 64.1 & 60.8 & 76.8 & 67.2 \\

Lingshu-32B~\citep{lingshu}
& Open
& 82.3 & 64.8 & 84.0 & 77.0 \\

\addlinespace[2pt]

Claude-Opus-5~\citep{claude_opus5}
& Closed
& 80.7 & 69.6 & 83.0 & 77.8 \\

GPT-5.6~\citep{gpt56}
& Closed
& 83.1 & \second{74.3} & 78.8 & 78.7 \\

Gemini-3.7-Flash~\citep{gemini37flash}
& Closed
& 83.0 & \second{74.3} & 83.4 & 80.2 \\

\midrule

CARE-Flow-S (4B)~\citep{care}
& Open
& 78.4 & 56.3 & 94.5 & 76.4 \\

CARE-Coord-S~\citep{care}
& Mixed
& 77.2 & 62.8 & \second{97.7} & 79.2 \\

CARE-Flow-B (10B)~\citep{care}
& Open
& 83.2 & 63.6 & 96.2 & \third{81.0} \\

CARE-Coord-B~\citep{care}
& Mixed
& 83.1 & 68.3 & \best{98.0} & \second{83.1} \\

Ophiuchus-7B~\citep{ophiuchus}
& Open
& \third{83.9} & 73.6 & 78.6 & 78.7 \\

\midrule

\rowcolor{oursyellow}
\textbf{GPT-5.6 + \method$^{\mathrm{C}}$}
& Mixed
& 83.7
& \third{73.8}
& 75.3
& 77.6 \\

\rowcolor{oursyellow}
\textbf{Gemini-3.7-Flash + \method$^{\mathrm{C}}$}
& Mixed
& \second{85.0}
& \best{74.7}
& 80.7
& 80.1 \\

\rowcolor{oursyellow}
\textbf{Qwen3.5-9B + \method}
& Open
& \best{89.9}
& 71.4
& \third{97.5}
& \best{86.3} \\

\bottomrule

\end{tabular}%
\end{adjustbox}

\end{table}

\definecolor{gainred}{RGB}{190,60,60}

\newcommand{\benchmarkscore}[2]{%
  \makebox[1.55em][r]{#1}%
  \makebox[1.45em][l]{%
    \textsuperscript{%
      \fontsize{5.5}{6}\selectfont\bfseries
      \textcolor{gainred}{#2}%
    }%
  }%
}

\begin{table}[!tb]
\centering
\hypersetup{hidelinks}

\caption{
\textbf{Evidence management versus direct tool use.}
Accuracy (\%) on seven medical VQA datasets, grouped by backbone.
Tool overlays a predicted mask and accepts the resulting answer directly.
Superscripts show percentage-point gains over the corresponding Tool row; yellow marks our methods.
\method$^{\mathrm{C}}$ uses a fixed overlay; full \method includes model fine-tuning and learned evidence presentation.
COVID denotes COVID-QU-Ex. Results use our original splits.
}
\label{tab_agentic_benchmark}

\compacttable
\footnotesize
\setlength{\tabcolsep}{2.5pt}
\renewcommand{\arraystretch}{1.14}

\begin{adjustbox}{width=\linewidth}
\begin{tabular}{lc*{7}{c}}

\toprule
\multicolumn{1}{c}{\multirow{2}{*}{\textbf{System}}}
& \multirow{2}{*}{\textbf{Harness}}
& \multicolumn{4}{c}{\textbf{Mask-Evidence MedVQA}}
& \multicolumn{3}{c}{\textbf{Public Medical VQA}} \\
\cmidrule(lr){3-6}\cmidrule(lr){7-9}
&
& BreastUS
& COVID
& NeoPolyp
& KiTS23
& SLAKE
& VQA-RAD
& OMVQA-3k \\
\midrule

GPT-5.6
& Tool
& \benchmarkscore{82.0}{}
& \benchmarkscore{63.5}{}
& \benchmarkscore{75.5}{}
& \benchmarkscore{72.9}{}
& \benchmarkscore{78.6}{}
& \benchmarkscore{71.6}{}
& \benchmarkscore{64.3}{} \\

\rowcolor{oursyellow}
& \textbf{\method}\rlap{$^{\mathrm{C}}$}
& \benchmarkscore{85.9}{+3.9}
& \benchmarkscore{66.6}{+3.1}
& \benchmarkscore{81.0}{+5.5}
& \benchmarkscore{73.7}{+0.8}
& \benchmarkscore{83.7}{+5.1}
& \benchmarkscore{73.8}{+2.2}
& \benchmarkscore{75.3}{+11.0} \\

\addlinespace[7pt]

Gemini-3.7-Flash
& Tool
& \benchmarkscore{83.6}{}
& \benchmarkscore{74.8}{}
& \benchmarkscore{72.5}{}
& \benchmarkscore{71.2}{}
& \benchmarkscore{80.0}{}
& \benchmarkscore{71.8}{}
& \benchmarkscore{75.0}{} \\

\rowcolor{oursyellow}
& \textbf{\method}\rlap{$^{\mathrm{C}}$}
& \benchmarkscore{84.9}{+1.3}
& \benchmarkscore{76.1}{+1.3}
& \benchmarkscore{75.0}{+2.5}
& \benchmarkscore{75.2}{+4.0}
& \benchmarkscore{85.0}{+5.0}
& \benchmarkscore{74.7}{+2.9}
& \benchmarkscore{80.7}{+5.7} \\

\addlinespace[7pt]

Qwen3.5-9B
& Tool
& \benchmarkscore{83.6}{}
& \benchmarkscore{62.7}{}
& \benchmarkscore{77.5}{}
& \benchmarkscore{73.0}{}
& \benchmarkscore{74.8}{}
& \benchmarkscore{67.4}{}
& \benchmarkscore{81.2}{} \\

\rowcolor{oursyellow}
& \textbf{\method}\rlap{$^{\mathrm{C}}$}
& \benchmarkscore{87.5}{+3.9}
& \benchmarkscore{65.3}{+2.6}
& \benchmarkscore{79.2}{+1.7}
& \benchmarkscore{76.8}{+3.8}
& \benchmarkscore{77.1}{+2.3}
& \benchmarkscore{70.3}{+2.9}
& \benchmarkscore{83.3}{+2.1} \\

\rowcolor{oursyellow}
& \textbf{\method}
& \benchmarkscore{89.8}{+6.2}
& \benchmarkscore{72.3}{+9.6}
& \benchmarkscore{83.0}{+5.5}
& \benchmarkscore{78.5}{+5.5}
& \benchmarkscore{90.2}{+15.4}
& \benchmarkscore{71.6}{+4.2}
& \benchmarkscore{98.2}{+17.0} \\

\bottomrule

\end{tabular}
\end{adjustbox}

\end{table}

\begin{figure}[t]
\centering
\includegraphics[width=0.9\linewidth]{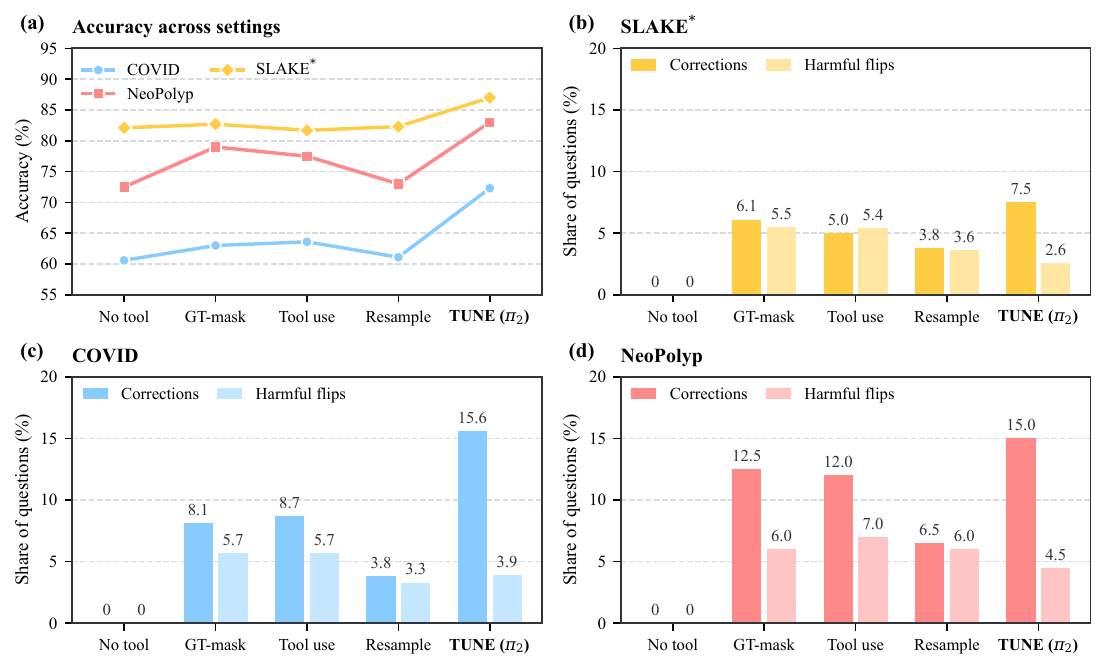}
\caption{\textbf{Mask evidence can correct or overturn no-tool answers.}
\textbf{(a)} Accuracy (\%) on COVID-QU-Ex, NeoPolyp, and SLAKE$^{*}$.
\textbf{(b--d)} Corrections (wrong$\to$right) and harmful flips (right$\to$wrong) are reported as percentages of all evaluated questions relative to the same SFT no-tool answers.
No tool, GT-mask, Tool use, and Resample use matched SFT weights. TUNE uses $\pi_2$. GT-mask and Tool use supply reference and predicted masks, respectively. For Resample, the SFT model selects between two answers it generates without tools.
Predicted-mask Dice is 0.973, 0.931, and 0.808 for COVID-QU-Ex, NeoPolyp, and SLAKE$^{*}$, respectively.}
\label{fig:main_comparison}
\end{figure}

\paragraph{Models and baselines.}
Full TUNE is built on Qwen3.5-9B~\citep{qwen35_9b}, which we fine-tune using the trajectories described in Section~\ref{sec_method_training}.
We additionally evaluate GPT-5.6 and Gemini-3.7-Flash~\citep{gpt56,gemini37flash} with \method$^{\mathrm{C}}$, which retains evidence planning and answer review but uses a fixed overlay interface. Their Critics use separate Qwen-based LoRA adapters.
For each backbone, we compare against direct tool use, which overlays the predicted mask and accepts the resulting answer directly.
We also compare with tool-free VLMs and prior medical agents where applicable.

\paragraph{Metrics.}
We report answer accuracy, \emph{corrections} (incorrect-to-correct changes), and \emph{harmful flips} (correct-to-incorrect changes). In Fig.~\ref{fig:main_comparison}, both rates are measured over all evaluated questions relative to the same SFT no-tool answers.
The latter two expose the information--bias trade-off that aggregate accuracy can hide. Closed answers use exact match after normalization; open answers use GPT-4o judging following CARE\textquotesingle{}s protocol.

\subsection{Main Results}
\label{sec:exp_main}

\paragraph{Does evidence management improve over direct tool use?}
Table~\ref{tab_agentic_benchmark} compares TUNE with direct tool use for each VLM backbone across all seven datasets.
Our evidence-management variants improve every backbone on every dataset.
For Qwen3.5-9B, full TUNE gains 9.1 percentage points on average over direct tool use.
Even with presentation fixed to an overlay, \method$^{\mathrm{C}}$ consistently improves GPT-5.6 and Gemini-3.7-Flash.
Table~\ref{tab_benchmark_context_vqa} provides broader context against existing medical VLMs and agents.
Qwen3.5-9B with TUNE reaches 86.3\% mean accuracy on CARE's released splits. With InternVL3-8B, TUNE reaches 82.0\%, compared with 81.0\% for CARE-Flow-B and 83.1\% for CARE-Coord-B (Appendix~\ref{app:care}). These results show competitive performance across backbones.

\paragraph{How often does a good mask help versus hurt?}
Fig.~\ref{fig:main_comparison} compares matched SFT controls with full TUNE ($\pi_2$). On COVID-QU-Ex, a ground-truth mask corrects answers on 8.1\% of questions and overturns initially correct answers on 5.7\%. Both ground-truth and predicted masks yield more corrections than Resample on all three evaluation sets. Compared with SFT direct tool use, TUNE yields more corrections on every set and a lower mean harmful-flip rate (3.7\% vs.\ 6.0\%).

\subsection{Ablations and Design Analysis}
\label{sec:exp_analysis}

\paragraph{Does evidence management help a fine-tuned backbone?}
With $\pi_2$ fixed, TUNE raises mean accuracy from 77.2\% to 83.4\% across seven datasets (Table~\ref{tab:training_controls}). Matched SFT and component-level comparisons are provided in Appendix~\ref{app:controls}.

\begin{table}[H]
\centering
\caption{\textbf{Fine-tuning and evidence-management gains.}
Accuracy (\%) on seven datasets using the original splits. Fine-tuning = $\pi_2$ (no tool) $-$ Base (no tool); evidence management = Full TUNE $-$ $\pi_2$ (no tool). Gains are in percentage points, computed from the displayed accuracies. COVID denotes COVID-QU-Ex.}
\label{tab:training_controls}
\compacttable
\setlength{\tabcolsep}{2.5pt}
\begin{adjustbox}{width=\linewidth}
\begin{tabular}{l*{8}{c}}
\toprule
\multicolumn{1}{c}{\multirow{2}{*}{\textbf{Configuration}}}
& \multicolumn{4}{c}{\textbf{Mask-Evidence MedVQA}}
& \multicolumn{3}{c}{\textbf{Public Medical VQA}}
& \multirow{2}{*}{\textbf{Mean}} \\
\cmidrule(lr){2-5}\cmidrule(lr){6-8}
& \multicolumn{1}{c}{BreastUS}
& \multicolumn{1}{c}{COVID}
& \multicolumn{1}{c}{NeoPolyp}
& \multicolumn{1}{c}{KiTS23}
& \multicolumn{1}{c}{SLAKE}
& \multicolumn{1}{c}{VQA-RAD}
& \multicolumn{1}{c}{OMVQA-3k}
& \\
\midrule
Base, no tool & 82.8 & 58.8 & 72.0 & 71.3 & 75.1 & 69.0 & 81.4 & 72.9 \\
$\pi_2$, no tool & 84.4 & 61.5 & 73.5 & 72.9 & 84.5 & 69.8 & 93.6 & 77.2 \\
\rowcolor{oursyellow}
\textbf{Full TUNE} & \textbf{89.8} & 72.3 & \textbf{83.0} & \textbf{78.5} & \textbf{90.2} & \textbf{71.6} & \textbf{98.2} & \textbf{83.4} \\
\midrule
Fine-tuning & +1.6 & \textbf{+2.7} & +1.5 & \textbf{+1.6} & \textbf{+9.4} & +0.8 & \textbf{+12.2} & +4.3 \\
\rowcolor{oursyellow}
\textbf{Evidence management} & +5.4 & \textbf{+10.8} & +9.5 & \textbf{+5.6} & +5.7 & \textbf{+1.8} & +4.6 & \textbf{+6.2} \\
\bottomrule
\end{tabular}
\end{adjustbox}
\end{table}

\paragraph{Can rendering alone change outcomes?}
Table~\ref{tab_interface_effect} holds the underlying mask fixed and changes only how it is presented.
The effects are large and bidirectional.
Relative to an overlay, calipers increase accuracy by 14.7 percentage points on COVID-QU-Ex but decrease it by 6.0 points on NeoPolyp.
A crop changes accuracy on the same datasets by $+9.7$ and $-7.5$ percentage points, respectively.
More broadly, no interface is consistently beneficial across image types.
Even with localization unchanged, presentation can alter what the VLM extracts from the evidence.
This supports treating interface selection as part of reasoning rather than a fixed visualization choice.

\begin{table}[H]
\centering
\caption{\textbf{Rendering effects vary across datasets.}
Changes in GPT-5.6 accuracy (percentage points) relative to the overlay interface,
with masks held fixed across renderings. Dashes mark the reference.
Mask-Evidence MedVQA uses ground-truth masks; public pools use predicted masks.
COVID denotes \mbox{COVID-QU-Ex}.}
\label{tab_interface_effect}
\compacttable
\setlength{\tabcolsep}{2.5pt}
\begin{adjustbox}{width=\linewidth}
\begin{tabular}{l*{7}{c}}
\toprule
\multicolumn{1}{c}{\multirow{2}{*}{\textbf{Interface}}}
& \multicolumn{4}{c}{\textbf{Mask-Evidence MedVQA}}
& \multicolumn{3}{c}{\textbf{Public pools}} \\
\cmidrule(lr){2-5}\cmidrule(lr){6-8}
& BreastUS
& COVID
& NeoPolyp
& KiTS23
& VQA-RAD
& OMVQA-3k
& MedXpertQA \\
\midrule

overlay (reference)
& \textemdash & \textemdash & \textemdash & \textemdash
& \textemdash & \textemdash & \textemdash \\

outline
& +0.8 & +6.0 & -3.5 & -0.3 & +0.9 & +2.0 & 0.0 \\

outline (cyan)
& +1.6 & +4.3 & -4.0 & +1.7 & -1.9 & 0.0 & +0.8 \\

contour fill
& +1.6 & +2.7 & -3.0 & -1.3 & +1.4 & -3.2 & +1.2 \\

bounding box
& -0.8 & +6.3 & -3.5 & -3.3 & -2.4 & -3.6 & +2.0 \\

circle
& -1.6 & +8.3 & -6.5 & +0.3 & -1.4 & -1.6 & -0.4 \\

crosshair
& -3.9 & +7.3 & -2.0 & -2.0 & -2.4 & +1.2 & -0.8 \\

arrow
& -3.1 & -2.0 & -5.5 & -2.3 & -1.4 & -2.4 & -0.8 \\

caliper
& -0.8 & +14.7 & -6.0 & 0.0 & 0.0 & -2.0 & -0.8 \\

crop
& -4.7 & +9.7 & -7.5 & -6.0 & -0.9 & -0.4 & -1.2 \\

inset
& -0.8 & +5.0 & -3.5 & +1.0 & +1.4 & -2.0 & -0.4 \\

mask only
& +0.8 & +3.3 & -1.5 & -3.7 & -0.9 & 0.0 & -0.8 \\

\bottomrule
\end{tabular}%
\end{adjustbox}
\end{table}

\begin{table}[!t]
\centering

\caption{\textbf{Component ablation of \method.}
Accuracy (\%) as target planning, learned evidence presentation, and answer selection
are progressively enabled. The Critic uses one-shot prompting or training on decisive answer pairs.
Bold marks the highest accuracy per dataset; yellow highlights full \method. Rows 1--2 use $\pi_0$, rows 3--4 use $\pi_1$, and row 5 uses $\pi_2$. COVID denotes COVID-QU-Ex.}
\label{tab_components_perdataset}

\compacttable
\begin{adjustbox}{max width=\linewidth}
\begin{tabular}{ccccccc}
\toprule
\textbf{Planner}
& \textbf{Presenter}
& \textbf{Critic}
& \textbf{BreastUS}
& \textbf{COVID}
& \textbf{NeoPolyp}
& \textbf{KiTS23} \\
\midrule

\textemdash & \textemdash & \textemdash
& 82.8 & 58.8 & 72.0 & 71.3 \\

\checkmark & \textemdash & \textemdash
& 83.6 & 62.7 & 77.5 & 73.0 \\

\checkmark & \checkmark & \textemdash
& 85.9 & 71.4 & 77.5 & 74.5 \\

\checkmark & \checkmark & \textit{One-shot}
& 85.2 & 70.3 & 77.0 & 71.3 \\

\rowcolor{oursyellow}
\checkmark & \checkmark & \textit{Trained}
& \textbf{89.8} & \textbf{72.3} & \textbf{83.0} & \textbf{78.5} \\

\bottomrule
\end{tabular}
\end{adjustbox}
\end{table}

\paragraph{Do learned presentation and answer review improve evidence use?}
Table~\ref{tab_components_perdataset} progressively enables the components of TUNE.
Adding the Target Planner improves over no-tool answering on all four Mask-Evidence MedVQA datasets.
The learned Presenter further improves BreastUS, COVID-QU-Ex, and KiTS23 and matches performance on NeoPolyp.
Answer review becomes beneficial only after training: a one-shot Critic lowers accuracy on all four datasets, whereas training on decisive pairs yields the highest accuracy on each.
These results show that both presentation and answer review benefit from supervision derived from downstream answer outcomes rather than being specified by prompting alone. The Presenter also outperforms the best fixed interface on three of four datasets and transfers more effectively within a modality than across modalities (Appendix~\ref{app:transfer}).

\paragraph{What information does the Critic need?}
We isolate the Critic on 27 decisive NeoPolyp pairs, each containing one correct and one incorrect candidate answer.
With answer text alone, the Critic selects the correct answer in 55.6\% of pairs.
Adding the original image raises selection accuracy to 63.0\%, while additionally providing the selected mask-derived rendering increases it to 74.1\%.
Thus, access to the evidence underlying an answer change improves selection accuracy by 18.5 percentage points over text alone, indicating that reliable answer review benefits from inspecting the visual evidence rather than comparing candidate answers in isolation. Additional Critic and mask-quality results are provided in Appendices~\ref{app:critic} and~\ref{app:masks}.

\section{Conclusion}

We introduced \textbf{TUNE}, an evidence-management harness for medical VLM agents.
Future work includes extending evidence management to richer medical tools and developing lighter, more scalable agent frameworks.

\FloatBarrier
\begingroup
\addtolength{\bibsep}{-0.8pt}

\endgroup

\clearpage
\appendix
\section{Benchmark Construction and Human Review}
\label{app:construction}
All questions are multiple choice with shuffled options. We exclude BUSI normal images, mixed-class NeoPolyp images, and KiTS23 slices containing both tumor and cyst or neither; KiTS23 is split by patient. Masks are binarized without class-coded colors. COVID-QU-Ex uses lung fields across all three classes; infection annotations are available only for COVID-positive images. The majority baseline always predicts the most frequent class.

\begin{table}[H]
\centering\compacttable
\caption{\textbf{Evaluation-set composition.} Majority and chance accuracies are percentages; chance assumes uniform selection among the listed classes.}
\label{tab:benchmark_statistics}
\setlength{\tabcolsep}{4pt}
\begin{tabular*}{0.98\linewidth}{@{\extracolsep{\fill}}lcccl@{}}
\toprule
\textbf{Dataset} & \textbf{$N$} & \textbf{Majority} & \textbf{Chance} & \textbf{Class composition} \\
\midrule
BreastUS & 128 & 67.2 & 50.0 & Benign 86; malignant 42 \\
COVID-QU-Ex & 1,166 & 50.0 & 33.3 & COVID 583; non-COVID 292; normal 291 \\
NeoPolyp & 200 & 72.0 & 50.0 & Neoplastic 144; non-neoplastic 56 \\
KiTS23 & 1,000 & 50.0 & 50.0 & Tumor 500; cyst 500 \\
\bottomrule
\end{tabular*}
\end{table}

Three reviewers assess 200 stratified examples. Fleiss\textquotesingle{} $\kappa=0.71$; criterion-level results are shown below. The final row requires all three criteria to be met.

\begin{table}[H]
\centering\compacttable
\caption{\textbf{Human review of 200 stratified examples.} Values are percentages.}
\label{tab:human_audit}
\begin{adjustbox}{max width=\linewidth}
\begin{tabular}{lcc}
\toprule
\textbf{Criterion} & \textbf{Majority says yes} & \textbf{All three say yes} \\
\midrule
Answerable from the original image alone & 97.5 & 95.0 \\
Consistent with the source label & 99.5 & 99.0 \\
Ground-truth mask relevant to the question & 98.0 & 95.0 \\
All three criteria & 96.0 & 90.5 \\
\bottomrule
\end{tabular}
\end{adjustbox}
\end{table}

\section{Training Data and Implementation}
\label{app:training}
$\mathcal D_{\mathrm{train}}$ contains 18,480 questions. Stage 1 randomly samples 60\% for interface trajectories; Stage 2 uses the full pool for decisive answer pairs. Evaluation questions are excluded. These counts refer to source questions before trajectory filtering. The shared model uses LoRA with a frozen vision encoder; the Planner has no dedicated target-selection supervision.

\begin{table}[H]
\centering\compacttable
\caption{\textbf{Training and test question counts.} CARE-aligned experiments use the released CARE splits.}
\label{tab:training_splits}
\begin{adjustbox}{max width=\linewidth}
\begin{tabular}{lcc}
\toprule
\textbf{Dataset} & \textbf{Training} & \textbf{Test} \\
\midrule
BreastUS & 517 & 128 \\
COVID-QU-Ex & 4,660 & 1,166 \\
NeoPolyp & 756 & 200 \\
KiTS23 & 1,835 & 1,000 \\
SLAKE & 4,919 & 1,061 \\
VQA-RAD & 1,793 & 451 \\
OmniMedVQA & 4,000 & 3,000 \\
\midrule
Total & 18,480 & 7,006 \\
\bottomrule
\end{tabular}
\end{adjustbox}
\end{table}

\begin{table}[H]
\centering\compacttable
\caption{\textbf{Fine-tuning settings.} Shared TUNE uses LoRA dropout 0.05 and 3\% warmup.}
\label{tab:adaptation_settings}
\begin{tabular*}{0.90\linewidth}{@{\extracolsep{\fill}}lcc@{}}
\toprule
\textbf{Setting} & \textbf{Shared TUNE} & \textbf{API-specific Qwen Critics} \\
\midrule
Adaptation & LoRA & Separate LoRA adapters \\
Vision encoder & Frozen & Frozen \\
Learning rate & $10^{-4}$ & $10^{-4}$ \\
Rank / alpha & 32 / 64 & 16 / 32 \\
Epochs & 2 & 3 \\
Effective batch size & 16 & 4 \\
Schedule & Cosine & Linear \\
\bottomrule
\end{tabular*}
\end{table}

\section{Training and Component Controls}
\label{app:controls}
Matched SFT uses gold answers to the same training questions and the same LoRA settings as TUNE. Component comparisons hold $\pi_2$ fixed. These controls separate model adaptation from the contributions of presentation and answer selection.

\begin{table}[H]
\centering\compacttable
\caption{\textbf{Matched SFT controls, accuracy (\%).} Direct tool use applies a fixed overlay without the Presenter or Critic.}
\label{tab:matched_sft}
\setlength{\tabcolsep}{4pt}
\begin{adjustbox}{max width=\linewidth}
\begin{tabular}{lccccc}
\toprule
\textbf{Dataset} & \textbf{SFT, no tool} & \textbf{SFT + tool} & \textbf{$\pi_2$, no tool} & \textbf{$\pi_2$ + tool} & \textbf{Full TUNE} \\
\midrule
SLAKE & 83.6 & 83.2 & 84.5 & 84.0 & 90.2 \\
OMVQA-3k & 92.8 & 92.5 & 93.6 & 93.2 & 98.2 \\
COVID-QU-Ex & 60.6 & 63.6 & 61.5 & 64.1 & 72.3 \\
NeoPolyp & 72.5 & 77.5 & 73.5 & 78.0 & 83.0 \\
\midrule
Mean & 77.4 & 79.2 & 78.3 & 79.8 & 85.9 \\
\bottomrule
\end{tabular}
\end{adjustbox}
\end{table}

\begin{table}[H]
\centering\compacttable
\caption{\textbf{Component accuracy (\%) at fixed $\pi_2$ weights.} The first row is the $\pi_0$ reference.}
\label{tab:fixed_weight_components}
\setlength{\tabcolsep}{4pt}
\begin{adjustbox}{max width=\linewidth}
\begin{tabular}{lccccc}
\toprule
\textbf{Weights and inference} & \textbf{SLAKE} & \textbf{OMVQA-3k} & \textbf{COVID-QU-Ex} & \textbf{NeoPolyp} & \textbf{Mean} \\
\midrule
$\pi_0$, no tool & 75.1 & 81.4 & 58.8 & 72.0 & 71.8 \\
$\pi_2$, no tool & 84.5 & 93.6 & 61.5 & 73.5 & 78.3 \\
$\pi_2$, direct overlay & 84.0 & 93.2 & 64.1 & 78.0 & 79.8 \\
$\pi_2$, overlay + Critic & 87.0 & 95.9 & 66.8 & 81.0 & 82.7 \\
$\pi_2$, Presenter, no Critic & 87.6 & 96.4 & 71.8 & 78.5 & 83.6 \\
$\pi_2$, full TUNE & 90.2 & 98.2 & 72.3 & 83.0 & 85.9 \\
\bottomrule
\end{tabular}
\end{adjustbox}
\end{table}

Pair-collection variants use the same questions and final training settings. The three-model configuration uses $\pi_0$ for planning and separate Presenter/answer and Critic models.

\begin{table}[H]
\centering\compacttable
\caption{\textbf{Harness used for collecting Critic supervision, accuracy (\%).}}
\label{tab:pair_collection}
\begin{adjustbox}{max width=\linewidth}
\begin{tabular}{lcccc}
\toprule
\textbf{Pair-collection harness} & \textbf{COVID-QU-Ex} & \textbf{NeoPolyp} & \textbf{SLAKE} & \textbf{Mean} \\
\midrule
Fixed overlay & 71.5 & 80.5 & 88.8 & 80.3 \\
Presenter-selected interfaces & 72.3 & 83.0 & 90.2 & 81.8 \\
\bottomrule
\end{tabular}
\end{adjustbox}
\end{table}

\begin{table}[H]
\centering\compacttable
\caption{\textbf{Shared versus role-specific models, accuracy (\%).}}
\label{tab:shared_roles}
\begin{adjustbox}{max width=\linewidth}
\begin{tabular}{lccccc}
\toprule
\textbf{Configuration} & \textbf{SLAKE} & \textbf{OMVQA-3k} & \textbf{COVID-QU-Ex} & \textbf{NeoPolyp} & \textbf{Mean} \\
\midrule
Three models & 90.4 & 98.0 & 72.6 & 83.5 & 86.1 \\
One shared model & 90.2 & 98.2 & 72.3 & 83.0 & 85.9 \\
\bottomrule
\end{tabular}
\end{adjustbox}
\end{table}

\section{Segmentation Quality and Target-Matched Controls}
\label{app:masks}
BiomedParse is evaluated on annotated targets in Table~\ref{tab:segmentation_quality}. Ground-glass-opacity predictions are compared with infection annotations on 583 COVID-positive images (median Dice 0.876). Wrong-mask and Dice-stratified comparisons use fixed $\pi_2$ weights. They assess sensitivity to misleading evidence and performance at different levels of localization quality. GT replacement is restricted to the 1,564 examples whose requested target matches an annotation.

\begin{table}[H]
\centering\compacttable
\caption{\textbf{Target-specific segmentation quality.} COVID ground-glass-opacity predictions use infection annotations as reference.}
\label{tab:segmentation_quality}
\begin{tabular*}{0.94\linewidth}{@{\extracolsep{\fill}}llcc@{}}
\toprule
\textbf{Dataset} & \textbf{Target / reference} & \textbf{$N$} & \textbf{Mean Dice} \\
\midrule
BreastUS & Breast lesion & 128 & 0.859 \\
COVID-QU-Ex & Lungs & 1,166 & 0.973 \\
COVID-QU-Ex & Ground-glass opacity / infection & 583 & 0.817 \\
NeoPolyp & Polyp & 200 & 0.931 \\
KiTS23 & Whole kidney & 1,000 & 0.897 \\
KiTS23 & Renal lesion (tumor or cyst) & 1,000 & 0.709 \\
\bottomrule
\end{tabular*}
\end{table}

\begin{table}[H]
\centering\compacttable
\caption{\textbf{Predicted and random wrong masks.} Accuracy (\%) is pooled over 2,494 questions.}
\label{tab:wrong_masks}
\begin{adjustbox}{max width=\linewidth}
\begin{tabular}{lccccc}
\toprule
\textbf{Mask} & \textbf{$N$} & \textbf{No tool} & \textbf{Direct tool} & \textbf{Presenter, no Critic} & \textbf{Full TUNE} \\
\midrule
Predicted & 2,494 & 68.2 & 70.2 & 74.5 & 76.5 \\
Wrong & 2,494 & 68.2 & 64.2 & 63.5 & 68.6 \\
\bottomrule
\end{tabular}
\end{adjustbox}
\end{table}

\begin{table}[H]
\centering\compacttable
\caption{\textbf{Accuracy (\%) by predicted-mask Dice on 1,564 target-matched examples.} Gain is relative to no tool within each stratum.}
\label{tab:dice_strata}
\setlength{\tabcolsep}{4pt}
\begin{adjustbox}{max width=\linewidth}
\begin{tabular}{ccccccc}
\toprule
\multicolumn{1}{l}{\textbf{Dice}} & \textbf{$N$} & \textbf{No tool} & \textbf{Direct tool} & \textbf{Presenter, no Critic} & \textbf{Full TUNE} & \textbf{Gain} \\
\midrule
$\geq0.9$ & 887 & 75.6 & 79.3 & 80.6 & 82.3 & +6.7 \\
$0.7\leq\mathrm{Dice}<0.9$ & 428 & 75.2 & 77.8 & 79.0 & 81.5 & +6.3 \\
$<0.7$ & 249 & 65.1 & 60.2 & 61.0 & 71.5 & +6.4 \\
\bottomrule
\end{tabular}
\end{adjustbox}
\end{table}

\begin{table}[H]
\centering\compacttable
\caption{\textbf{Target-matched GT replacement, accuracy (\%).} The All row pools the retained examples.}
\label{tab:matched_gt}
\setlength{\tabcolsep}{4pt}
\begin{adjustbox}{max width=\linewidth}
\begin{tabular}{lccccc}
\toprule
\textbf{Requested / annotated target} & \textbf{$N$} & \textbf{Predicted Dice} & \textbf{TUNE, predicted} & \textbf{TUNE, GT} & \textbf{Change} \\
\midrule
BreastUS lesion & 123 & 0.86 & 90.2 & 89.4 & $-0.8$ \\
COVID-QU-Ex lungs & 359 & 0.97 & 78.8 & 78.6 & $-0.2$ \\
NeoPolyp polyp & 190 & 0.93 & 83.7 & 84.2 & $+0.5$ \\
KiTS23 whole kidney & 385 & 0.89 & 76.4 & 76.6 & $+0.2$ \\
KiTS23 renal lesion & 507 & 0.71 & 80.9 & 82.1 & $+1.2$ \\
\midrule
All & 1,564 & --- & 80.4 & 80.8 & $+0.4$ \\
\bottomrule
\end{tabular}
\end{adjustbox}
\end{table}

\section{Presentation Selection and Transfer}
\label{app:transfer}
All comparisons use $\pi_2$ without the Critic. The Presenter selects an interface per question, whereas each fixed-interface baseline uses one view throughout. Chest X-ray questions come from SLAKE and VQA-RAD; abdominal CT questions come from SLAKE, VQA-RAD, and OMVQA-3k.

\begin{table}[H]
\centering\compacttable
\caption{\textbf{Learned presentation versus a fixed interface, accuracy (\%).} The best fixed interface is selected with test labels.}
\label{tab:best_interface}
\begin{adjustbox}{max width=\linewidth}
\begin{tabular}{lcclc}
\toprule
\textbf{Dataset} & \textbf{Overlay} & \textbf{Best fixed} & \textbf{Interface} & \textbf{Presenter} \\
\midrule
BreastUS & 85.2 & 87.5 & Contour fill & 86.7 \\
COVID-QU-Ex & 64.1 & 70.4 & Caliper & 71.8 \\
NeoPolyp & 78.0 & 78.0 & Overlay & 78.5 \\
KiTS23 & 74.0 & 74.6 & Inset & 75.2 \\
\bottomrule
\end{tabular}
\end{adjustbox}
\end{table}

COVID-QU-Ex interface shares are 38\% caliper, 26\% crop, 17\% overlay, 11\% outline, and 8\% others.

\begin{table}[H]
\centering\compacttable
\caption{\textbf{Cross-dataset presentation transfer, accuracy (\%).} Best fixed is selected with test labels.}
\label{tab:interface_transfer}
\setlength{\tabcolsep}{4pt}
\begin{tabular*}{0.98\linewidth}{@{\extracolsep{\fill}}lllccc@{}}
\toprule
\textbf{Setting} & \multicolumn{1}{c}{\textbf{Training data}} & \textbf{Test data} & \textbf{Overlay} & \textbf{Best fixed} & \textbf{Presenter} \\
\midrule
Across modalities & BreastUS, NeoPolyp, KiTS23 & COVID-QU-Ex & 63.4 & 70.0 & 65.6 \\
Within modality & COVID-QU-Ex & Chest X-ray & 68.8 & 72.7 & 71.6 \\
Within modality & Abdominal CT & KiTS23 & 73.4 & 74.3 & 74.0 \\
\bottomrule
\end{tabular*}
\end{table}

\section{Critic Behavior and Resampling}
\label{app:critic}
A decisive pair contains exactly one correct candidate. For non-decisive pairs, either selection has the same correctness. The Critic runs when answers differ; candidate positions are randomized. Non-decisive disagreements account for 2.1\% of questions, and first-position selection is 50.3\% across disagreements.

\begin{table}[H]
\centering\compacttable
\caption{\textbf{Critic pair statistics (\%).} Share and both-wrong rates use all questions. Accuracy and tool-selection rates are conditional on pair type; non-decisive tool selection uses only Critic calls. N/A indicates no such calls.}
\label{tab:critic_pairs}
\setlength{\tabcolsep}{4pt}
\begin{adjustbox}{max width=\linewidth}
\begin{tabular}{lcccccc}
\toprule
& \multicolumn{3}{c}{\textbf{Decisive}} & \multicolumn{3}{c}{\textbf{Non-decisive}} \\
\cmidrule(lr){2-4}\cmidrule(lr){5-7}
\textbf{Dataset} & \textbf{Share} & \textbf{Accuracy} & \textbf{Tool chosen} & \textbf{Share} & \textbf{Both wrong} & \textbf{Tool chosen} \\
\midrule
SLAKE & 13.6 & 80.6 & 68.1 & 86.4 & 7.2 & 65.9 \\
VQA-RAD & 13.5 & 62.3 & 59.0 & 86.5 & 23.3 & 65.1 \\
OMVQA-3k & 7.3 & 93.6 & 69.1 & 92.7 & 1.3 & 66.7 \\
BreastUS & 16.4 & 76.2 & 61.9 & 83.6 & 6.3 & N/A \\
COVID-QU-Ex & 22.0 & 75.8 & 80.5 & 78.0 & 22.4 & 71.9 \\
NeoPolyp & 29.0 & 74.1 & 63.8 & 71.0 & 9.5 & N/A \\
KiTS23 & 16.9 & 76.3 & 56.8 & 83.1 & 17.5 & N/A \\
\midrule
All seven & 13.3 & 79.9 & 68.7 & 86.7 & 9.8 & 68.0 \\
\bottomrule
\end{tabular}
\end{adjustbox}
\end{table}

\begin{table}[H]
\centering\compacttable
\caption{\textbf{Corrections / harmful flips (\%) against SFT no-tool answers.} Rates use all evaluated questions. For Resample, the SFT model selects between two answers it generates without tools.}
\label{tab:resampling}
\begin{adjustbox}{max width=\linewidth}
\begin{tabular}{lccc}
\toprule
\textbf{Condition} & \textbf{COVID-QU-Ex} & \textbf{NeoPolyp} & \textbf{SLAKE*} \\
\midrule
GT-mask (SFT) & 8.1 / 5.7 & 12.5 / 6.0 & 6.1 / 5.5 \\
Tool use (SFT) & 8.7 / 5.7 & 12.0 / 7.0 & 5.0 / 5.4 \\
Resample (SFT) & 3.8 / 3.3 & 6.5 / 6.0 & 3.8 / 3.6 \\
TUNE ($\pi_2$) & 15.6 / 3.9 & 15.0 / 4.5 & 7.5 / 2.6 \\
\bottomrule
\end{tabular}
\end{adjustbox}
\end{table}

\section{CARE-Aligned Comparisons}
\label{app:care}
TUNE uses CARE's released splits and GPT-4o scoring. The InternVL3 version uses CARE's released training questions and the same answering backbone as CARE-Flow-B.

\begin{table}[H]
\centering\compacttable
\caption{\textbf{CARE-aligned accuracy (\%).} CARE and its InternVL3 SFT baseline are reported results. Segmentation tools and training budgets differ.}
\label{tab:care_aligned}
\setlength{\tabcolsep}{4pt}
\begin{adjustbox}{max width=\linewidth}
\begin{tabular}{lcccc}
\toprule
\textbf{System} & \textbf{SLAKE} & \textbf{VQA-RAD} & \textbf{OMVQA-3k} & \textbf{Mean} \\
\midrule
InternVL3-8B SFT & 76.5 & 61.9 & 91.1 & 76.5 \\
CARE-Flow-B & 83.2 & 63.6 & 96.2 & 81.0 \\
CARE-Coord-B, GPT-5 coordinator & 83.1 & 68.3 & 98.0 & 83.1 \\
TUNE, InternVL3-8B & 84.5 & 65.9 & 95.6 & 82.0 \\
Qwen3.5-9B SFT & 83.4 & 68.5 & 92.3 & 81.4 \\
TUNE, Qwen3.5-9B & 89.9 & 71.4 & 97.5 & 86.3 \\
\bottomrule
\end{tabular}
\end{adjustbox}
\end{table}

TUNE with InternVL3-8B obtains 79.4\% mean accuracy without the Critic, compared with 82.0\% with it.

\section{Agent Training Setups}
\label{app:agent_training}

Table~\ref{tab:agent_training} summarizes which components were trained and what supervision they received for the agents in Table~\ref{tab_benchmark_context_vqa}.
CARE-Flow evaluates a fixed set of evidence views and combines the resulting answers.
CARE-Coord uses the same trained specialists and adds a prompted GPT-5 coordinator that selects evidence views and reviews the reasoning and answers.

\begin{table}[H]
\centering\compacttable
\caption{\textbf{Training setups of the medical VQA agents compared in Table~\ref{tab_benchmark_context_vqa}.}}
\label{tab:agent_training}
\setlength{\tabcolsep}{4pt}
\renewcommand{\arraystretch}{1.12}
\begin{adjustbox}{max width=\linewidth}
\begin{tabular}{@{}>{\raggedright\arraybackslash}p{0.25\linewidth}>{\raggedright\arraybackslash}p{0.25\linewidth}>{\raggedright\arraybackslash}p{0.45\linewidth}@{}}
\toprule
\textbf{System} & \textbf{Components trained} & \textbf{Training supervision} \\
\midrule
CARE-Flow / CARE-Coord\newline\citep{care}
& Entity proposer, segmenter, and answering VLM
& RL for entity proposal, mask supervision for segmentation, and SFT followed by RL on in-domain VQA data for the answering VLM \\
\addlinespace[3pt]
Ophiuchus-7B\newline\citep{ophiuchus}
& Qwen2.5-VL-7B agent
& Cold-start SFT on medical tool-use trajectories, self-reflection fine-tuning, and agentic tool RL \\
\addlinespace[3pt]
TUNE (ours)
& Shared Qwen3.5-9B
& Supervised fine-tuning on successful presentation trajectories and decisive answer pairs \\
\addlinespace[3pt]
TUNE$^{\mathrm C}$ (ours)
& Qwen-based Critic
& Supervised fine-tuning on decisive answer pairs, with the API backbone kept fixed \\
\bottomrule
\end{tabular}
\end{adjustbox}
\end{table}

\clearpage
\begingroup
\raggedbottom
\section{Prompts}
\label{app:prompts}
This appendix provides the inference prompts for \method (Section~\ref{sec_method}) and the system template for synthesizing training trajectories.
Each inference block shows the system instructions followed by the inputs for a given example.
The Target Planner selects a region to segment, and the Evidence Presenter chooses how to display the mask using its geometry, the image, and the question.
The two answering prompts share a final-answer format. The evidence-conditioned prompt additionally supplies the rendered mask and asks the model to check that it matches the target before using it.
When the answers differ, the Answer Critic compares the two candidates.
For MedSAM, a separate prompt proposes a bounding box for the selected target.
\lstset{basicstyle=\ttfamily\fontsize{7}{8}\selectfont,breaklines=true,breakatwhitespace=true,breakautoindent=false,breakindent=0pt,columns=fullflexible,keepspaces=true,showstringspaces=false,frame=single,framerule=0.25pt,rulecolor=\color{black!20},backgroundcolor=\color{black!2},framesep=4pt,xleftmargin=5pt,xrightmargin=5pt,aboveskip=2pt,belowskip=6pt}

\par\noindent
\begin{minipage}{\linewidth}
\subsection{Target Planner}

\begin{lstlisting}
You are an expert radiologist answering a question about a medical image. Before you answer, a segmentation tool will isolate one target that you choose and draw it back onto the image for you. In this step your only job is to choose that target.

**Discriminative target selection.** A target that is merely relevant to the question is often too broad to provide useful evidence for reasoning. The organ or region the question is about is relevant in every image of this kind, so segmenting it shows you where to look but cannot tell the candidate answers apart. Prefer the finding whose presence, absence, extent, shape, margin or internal appearance would differ between the candidate answers -- a lesion, an abnormal region, or the specific structure that carries the distinguishing feature. Test your choice before you commit: if the mask would look the same whichever candidate answer is correct, it is not discriminative, so choose again.

Never name one of the candidate answers as the target. If the options are, say, "benign" and "malignant", ask for the lesion, not for "malignant tumor": the mask must be evidence you interpret, not a label that answers the question for you.

**Tool-aware planning.** Segmentation models differ both in what they can segment and in the inputs they require. The target you name is passed to the tool configured for this run:

- **BiomedParse** accepts text prompts for a predefined vocabulary of anatomical structures and lesions -- organs such as `lung`, `liver`, `kidney`, `spleen`, `pancreas`, cardiac chambers, and lesion-level terms such as `tumor`, `nodule`, `polyp`, `lesion`, `edema` -- but cannot segment targets outside that vocabulary. A phrase it does not know is mapped to the nearest term it does know, which is often the whole organ. As used here, it almost never returns an empty mask, so a mask is evidence that *something* was segmented, not that the requested finding is present.
- **MedSAM** accepts no text. After you name the target, a separate step draws a bounding box around it, and MedSAM segments the object inside that box. It can segment any visible object with a clear boundary, but only one that can be boxed: a diffuse pattern, a bilateral or scattered finding, or a finding that is absent cannot be localised this way.

So the target must be both discriminative and segmentable: the most specific term that names a single, localisable region the tool can act on -- for example `kidney lesion` rather than `abdomen`, `breast lesion` rather than `breast`. Fall back to the organ only when no more specific region can be named.

Work in two steps.

**First**, in at most three short lines, write the differential: the candidate answers, and the one target that separates them, checked against both rules above.

**Then** emit exactly one fenced call for that target and stop:

```tool
segment(target="<target>")
```

Lower case, one target, no answer option. Do not answer the question in these lines. You will be shown the segmentation and asked for your answer afterwards.

![]({{ image }}) {{ question }}

Do not answer this question yet. Write the differential, then emit the tool call.
\end{lstlisting}
\end{minipage}
\par\medskip

\par\noindent
\begin{minipage}{\linewidth}
\subsection{Evidence Presenter}

\begin{lstlisting}
You are choosing how a segmentation result should be drawn back onto a medical image before you are asked to answer a question about it.

You are shown the original image, the question, and the geometry of the mask the tool produced. You are NOT shown the mask drawn on the image -- the point of the choice is to make it before anything is rendered.

The available presentations are:

- `overlay` -- a translucent colour wash filling the masked region
- `outline` -- the region's contour only, the image otherwise untouched
- `outline_cyan` -- the same contour, drawn in cyan
- `contour_fill` -- the region's contour with the area inside it filled in
- `bbox` -- an axis-aligned rectangle around the region
- `circle` -- a circle drawn around the region
- `crosshair` -- a crosshair centred on the region's centroid
- `arrow` -- an arrowhead pointing at the region's centroid
- `caliper` -- paired caliper marks at the two ends of the region's longest axis
- `crop` -- the masked region alone, enlarged, without the surrounding anatomy
- `inset` -- the region enlarged into a panel placed beside the unmodified image
- `mask_only` -- the mask on its own, without the underlying image

They differ in how much of the surrounding anatomy they occlude, how precisely they delimit the boundary, and what they imply about the task -- a caliper says *measure this*, an arrow says *look here*, a crop removes the context entirely.

Reply with one presentation name and nothing else.

![]({{ image }})

Question that will be asked about this image: {{ question }}

Mask geometry: {{ stats }}

Which presentation gives the best chance of a correct answer? Reply with the name alone.
\end{lstlisting}
\end{minipage}
\par\medskip

\par\noindent
\begin{minipage}{\linewidth}
\subsection{Image-only answer}

\begin{lstlisting}
You are a doctor's assistant answering a question about a medical image.

Look at the image carefully and reason as much as you need to. When you are done, give your final answer on the last line, in exactly this form:

Answer: <text>

For a multiple-choice question `<text>` is the option letter alone, e.g. `Answer: B`. For a yes/no question it is `Answer: yes` or `Answer: no`. Otherwise it is the shortest phrase that fully answers the question.

![]({{ image }}) {{ question }}

End your reply with a final line in exactly this form: `Answer: <text>`
\end{lstlisting}
\end{minipage}
\par\medskip

\par\noindent
\begin{minipage}{\linewidth}
\subsection{Evidence-conditioned answer}

\begin{lstlisting}
You are a doctor's assistant answering a question about a medical image.

You are also shown the output of a segmentation tool run on that image. Look at both carefully and reason as much as you need to. When you are done, give your final answer on the last line, in exactly this form:

Answer: <text>

For a multiple-choice question `<text>` is the option letter alone, e.g. `Answer: B`. For a yes/no question it is `Answer: yes` or `Answer: no`. Otherwise it is the shortest phrase that fully answers the question.

![]({{ image }}) {{ question }}

![]({{ rendering }})

Above is the result of segmenting `{{ target }}`, drawn with the `{{ interface }}` presentation. {{ stats }}

Treat it as evidence, not as an answer. The tool can be wrong: it may have segmented the wrong structure, missed part of one, or found nothing meaningful. Judge from the image whether the highlighted region really is `{{ target }}` before you rely on it. If it is wrong, ignore it and answer from the original image.

Now answer the question: {{ question }}

End your reply with a final line in exactly this form: `Answer: <text>`
\end{lstlisting}
\end{minipage}
\par\medskip

\par\noindent
\begin{minipage}{\linewidth}
\subsection{Answer Critic}

\begin{lstlisting}
You are choosing which of two candidate answers to a medical imaging question is correct.

Both were produced by the same model. One saw the image alone; the other also saw a region that a segmentation tool proposed, and the text below says which is which. **Which answer sits in slot A is randomised per question**, so the letter itself carries no information -- choosing by letter scores at chance.

The two answers differ. They may still both be right, or both be wrong. Say which is more likely to be correct; do not treat the difference between them as necessarily the difference between right and wrong.

Judge only which answer is most likely **correct for the question asked**. Ignore length, confidence, fluency and hedging -- a terse right answer beats a thorough wrong one. An answer that narrates the mask in detail is not thereby more likely to be right.

Reply with exactly one character -- `A` or `B` -- and nothing else.

![]({{ image }})

Question: {{ question }}

The segmentation tool was asked to find `{{ target }}` and returned `{{ stats }}`. {{ rendering_note }}

![]({{ renderings }})

Answer A: {{ answer_a }}

Answer B: {{ answer_b }}

Which answer is correct? Reply `A` or `B`.
\end{lstlisting}
\end{minipage}
\par\medskip

\par\noindent
\begin{minipage}{\linewidth}
\subsection{Training-synthesis system template}

\begin{lstlisting}
You are a medical imaging analysis assistant. You answer clinical questions about {modality} images of the {body_part} by combining direct visual inspection with targeted segmentation-based inspection.

Your workflow should resemble visual exploration:

1. First inspect the original image directly.
2. Describe the visible findings relevant to the clinical question.
3. If a specific abnormality, structure, or region needs closer inspection, identify it as the target for segmentation.
4. Call the segmentation tool using <mask> to obtain a focused view of that target.
5. After the mask is provided, interpret what the highlighted region shows: where it is, whether it matches the suspected image finding, and what it reveals about the target's extent, morphology, or location.
6. Use the image evidence and segmentation result together to answer the clinical question.

Important:
- Segmentation is a focused visual inspection tool, similar to zooming into or isolating a clinically relevant region.
- Do not treat the mask as an oracle or replacement for image inspection.
- Do not state the final diagnosis before interpreting the mask.
- If one target is sufficient, use one segmentation step. If multiple targets are clinically relevant, inspect them one at a time.
\end{lstlisting}
\end{minipage}
\par\medskip

\par\noindent
\begin{minipage}{\linewidth}
\subsection{MedSAM bounding-box proposal}

\begin{lstlisting}
You are an expert radiologist examining a medical image. You have a segmentation tool that isolates a region and shows it back to you, and it must be given a bounding box.

You are told which structure to segment. Do not choose your own and do not judge whether it is the right one -- your only job is to locate the named structure in this image.

Emit exactly one fenced call and nothing else:

```tool
medsam(target="{{ target }}", box=[120, 45, 310, 260])
```

`box` is `[x0, y0, x1, y1]` in pixels, origin at the **top-left**, tightly enclosing `{{ target }}`.

The image is {{ width }} pixels wide and {{ height }} pixels tall, so every coordinate must fall inside those bounds and `x0 < x1`, `y0 < y1`. Judge the position from the image itself; do not guess a generic central box, because a box in the wrong place produces a segmentation of the wrong thing and is worse than no tool call at all.

If `{{ target }}` names something that has no localizable region in this image -- an imaging modality, a negation, a stage or grade -- box the anatomy the question is about instead.

![]({{ image }})

Question: {{ question }}

Locate: {{ target }}

Emit only the tool call.
\end{lstlisting}
\end{minipage}
\par\medskip

\endgroup

\end{document}